\documentclass[a4paper, 10pt, conference]{ieeeconf}      
\usepackage{FG2021}

\FGfinalcopy 

\IEEEoverridecommandlockouts                              
\usepackage[accsupp]{axessibility} 

\IEEEoverridecommandlockouts                             
\usepackage{array}
\usepackage{multicol}
\usepackage{tabularx, booktabs, ragged2e}
\usepackage[acronym]{glossaries}
\usepackage{acronym}
\usepackage{adjustbox}
\usepackage{multirow}
\usepackage{multicol}
\usepackage{balance}
\usepackage[colorlinks, linkcolor=red, urlcolor=blue]{hyperref}
\usepackage{amssymb}
\usepackage{blindtext}
\usepackage[acronym]{glossaries}
\usepackage{ltablex, makecell, float, textcomp}
\usepackage{colortbl}
\usepackage{bibunits}
\usepackage{comment}
\usepackage{pifont}
\usepackage{amsmath}
\usepackage{graphicx}
\usepackage{subcaption}
\usepackage{cite}
\usepackage{balance}
\usepackage{booktabs}

\usepackage{multirow}
\usepackage{xcolor}
\definecolor{diff}{rgb}{0.0, 0.0, 0.0}

\newcommand{\diff}[1]{\color{diff}{#1}~\color{black}}

\usepackage[utf8]{inputenc}

\DeclareFixedFont{\ttb}{T1}{txtt}{bx}{n}{7} 
\DeclareFixedFont{\ttm}{T1}{txtt}{m}{n}{7}  

\usepackage{color}
\definecolor{deepblue}{rgb}{0,0,0.5}
\definecolor{deepred}{rgb}{0.6,0,0}
\definecolor{deepgreen}{rgb}{0,0.5,0}
\definecolor{mygreen}{rgb}{0,0.6,0}
\definecolor{mygray}{rgb}{0.5,0.5,0.5}
\definecolor{mymauve}{rgb}{0.58,0,0.82}
\usepackage{listings}

\newcommand\pythonstyle{\lstset{
language=Python,
basicstyle=\ttm,
morekeywords={self, split},              
keywordstyle=\ttb\color{deepblue},
emph={MyClass,__init__},          
emphstyle=\ttb\color{deepred},    
stringstyle=\color{deepgreen},
frame=tb,                         
commentstyle=\color{mygreen},    
showstringspaces=true,
}}

\lstnewenvironment{python}[1][]
{
\pythonstyle
\lstset{#1}
}
{}

\newcommand{\etal}{\textit{et al}. }

\acrodef{fg}[FG]{16$^{th}$ IEEE International Conference on Automatic Face and Gesture Recognition}

\newcommand{\ie}{\textit{i}.\textit{e}., }
\newcommand{\eg}{\textit{e}.\textit{g}., }
\acrodef{gnn}[GNN]{graphical neural network}

\acrodef{ml}[ML]{machine learning}
\acrodef{sota}[SOTA]{state-of-the-art}

\acrodef{hog}[HOG]{histogram of gradients}

\acrodef{ss}[SS]{sister-sister}
\acrodef{bb}[BB]{brother-brother}
\acrodef{sibs}[SIBS]{brother-sister}

\acrodef{fs}[FS]{father-son}
\acrodef{ms}[MS]{mother-son}
\acrodef{fd}[FD]{father-daughter}

\acrodef{md}[MD]{mother-daughter}
\acrodef{gfgs}[GFGS]{grandfather-grandson}
\acrodef{gmgs}[GMGS]{grandmother-grandson}
\acrodef{gfgd}[GFGD]{grandfather-granddaughter}

\acrodef{gmgd}[GMGD]{grandmother-granddaughter}

\acrodef{sdm}[SDM]{signal detection model}
\acrodef{roc}[ROC]{receiver operating characteristic}
\acrodef{nmse}[NMSE]{Normalized Mean Square Error}
\acrodef{det}[DET]{Detection Error Trade-off}
\acrodef{tp}[TP]{true-positive}
\acrodef{tn}[TN]{true-negative}
\acrodef{ap}[AP]{average precision}

\acrodef{tpir}[TPIR]{true-positive identification rate}
\acrodef{frir}[FRIR]{false-reject identification rate}
\acrodef{fpir}[FRIR]{false-positive identification rate}

\acrodef{fn}[FN]{false-negative}
\acrodef{frr}[FRR]{false-reject rate}
\acrodef{fnr}[FNR]{false-negative rate}

\acrodef{fpr}[FPR]{false-positive rate}
\acrodef{tpr}[TPR]{true-positive rate}

\acrodef{bfw}[BFW]{\textit{Balanced Faces In the Wild}}

\acrodef{fiw}[FIW]{\textit{Families In the Wild}}
\acrodef{tsk}[TSKIN]{\textit{Tri-Subject Kinship}}
\acrodef{fiwmm}[FIW MM]{\textit{FIW in Multimedia}}

\acrodef{kfw}[KinFaceW]{\textit{Kin-Faces in the Wild}}
\acrodef{kfvw}[KFVW]{\textit{KinFaceW Videos}}

\acrodef{rfiw}[RFIW]{\textit{Recognizing Families In the Wild}}

\acrodef{cnn}[CNN]{Convolutional Neural Network}
\acrodef{lut}[LUT]{Look-Up-Table}
\acrodef{fr}[FR]{face recognition}

\acrodef{map}[mAP]{mean average precision}

\acrodef{mlp}[MLP]{multi-layer perceptron}

\acrodef{soa}[SOTA]{state-of-the-art}

\acrodef{svm}[SVM]{Support Vector Machine}
\acrodef{mid}[MID]{Member ID}
\acrodef{fid}[FID]{Family ID}
\acrodef{pid}[PID]{Photo ID}
\acrodef{roc}[ROC]{receiver operating characteristic}
\acrodef{nrml}[NRML]{Neighborhood Repulsed Metric Learning}

\definecolor{ao(english)}{rgb}{0.0, 0.5, 0.0}
	
\definecolor{diffy}{rgb}{0.0, 0.0, 1.0}
\definecolor{black}{rgb}{0.0, 0.0, 0.0}

\usepackage{colortbl}

\newcommand*\rot{\rotatebox{0}}

\title{\LARGE \bf
The 5th Recognizing Families in the Wild Data Challenge: Predicting Kinship from Faces
}
\definecolor{lightgray}{gray}{0.9}

\author{
    \parbox{16cm}{\centering
    \large Joseph P. Robinson$^{1,2}$, Can Qin$^2$, Ming Shao$^3$, Matthew A. Turk$^4$,\\Rama Chellappa$^5$, and Yun Fu$^2$\\
    \normalsize
    $^1$Vicarious Surgical $^2$Northeastern University $^3$UMass Dartmouth\\$^4$Toyota Technological Institute at Chicago (TTIC) $^5$Johns Hopkins University
}
}

\begin{document}

\IEEEoverridecommandlockouts\pubid{\makebox[\columnwidth]{978-1-6654-3176-7/21/\$31.00~\copyright{}2021 IEEE \hfill} \hspace{\columnsep}\makebox[\columnwidth]{ }}

\ifFGfinal
\thispagestyle{empty}
\pagestyle{empty}
\else
\author{Anonymous FG2021 submission\\ Paper ID \FGPaperID \\}
\pagestyle{plain}
\fi
\maketitle


\acresetall

\begin{abstract}
\ac{rfiw}, held as a data challenge in conjunction with the \ac{fg}, is a large-scale, multi-track visual kinship recognition evaluation. For the fifth edition of \ac{rfiw}, we continue to attract scholars, bring together professionals, publish new work, and discuss prospects. In this paper, we summarize submissions for the three tasks of this year’s \ac{rfiw}: specifically, we review the results for kinship verification, tri-subject verification, and family member search and retrieval. We look at the \ac{rfiw} problem, share current efforts, and make recommendations for promising future directions.
\end{abstract}
\hypersetup{citecolor=blue}
\acresetall
\glsresetall
\section{Introduction}
Automatic kinship recognition could be used for various uses, such as forensic analysis, automated photos app management, historical genealogy, multimedia analysis, missing kids and human trafficking tragedies, and immigration and border patrol concerns. Nonetheless, the challenges in such face-based tasks (\ie fine-grained classification in unconstrained settings) are only magnified in kin-based problem sets, as the data shows a high degree of variability in pose, illumination, background, and clarity, and soft biometric target labels, which only worsens the challenges with directional relationships consideration. As a result, the practical benefits of improving kinship-based technologies are counterbalanced by the difficulties posed by the problem of automated kinship comprehension. The \ac{rfiw} challenge series was born out of this need: a large-scale data challenge with various tasks to advance kinship detection technology. \ac{rfiw} is an open forum for researchers to present and discuss the \ac{soa}.

\begin{figure}[!t]
\begin{center}

  \includegraphics[width=\linewidth]{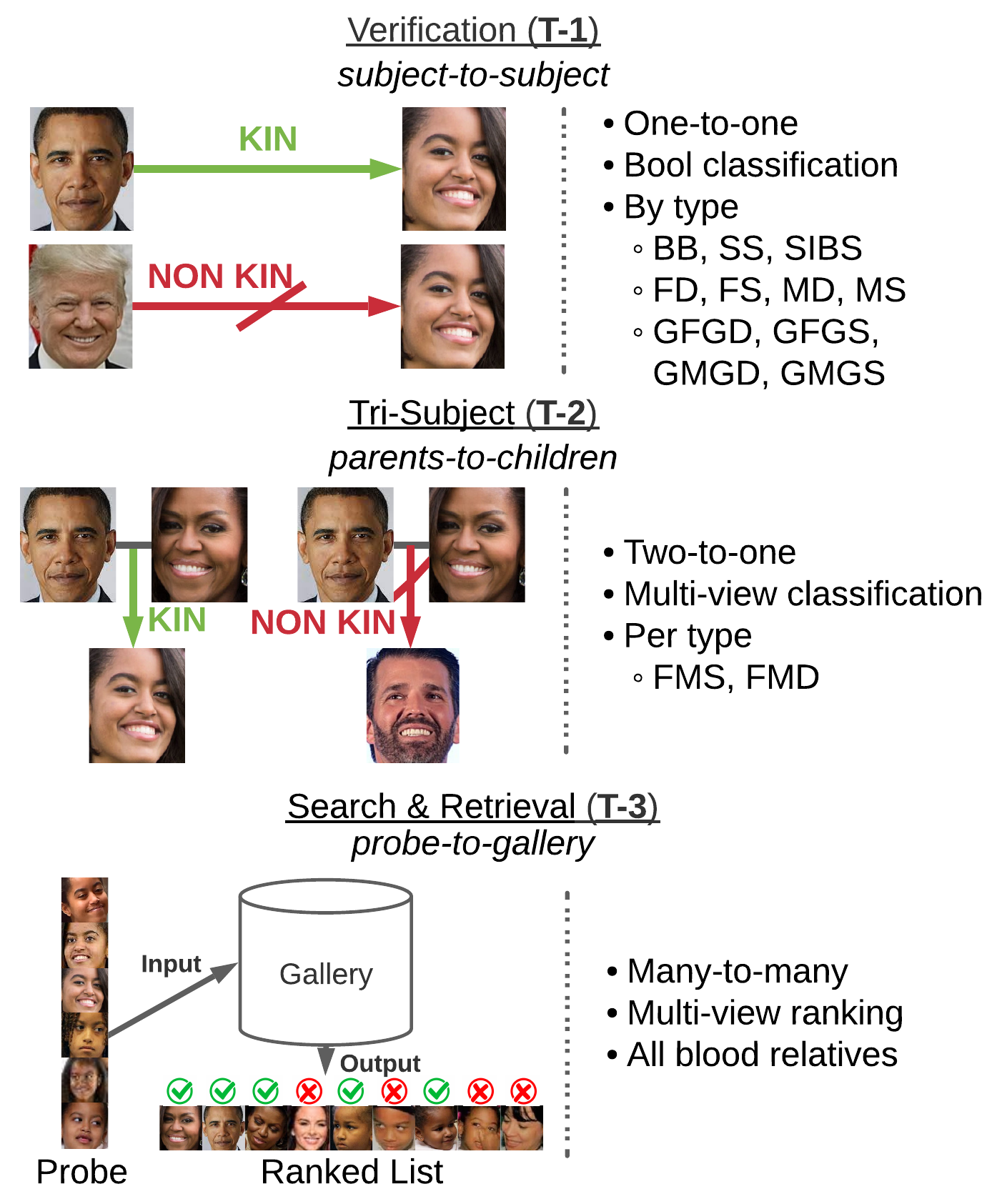}  
  \end{center}
\caption{An illustration of the three tasks supported in \ac{rfiw}.}
\label{fig:tasks:overview}
\end{figure}

In conjunction with the \ac{fg}, participants of the fifth \ac{rfiw}\footnote{\ac{rfiw}2021 webpage, \href{https://medium.com/to-recognize-families-in-the-wild-a-machine-vision/rfiw2021-7ceb357a39a6}{https://medium.com/recognize-families}.} 
 continue to push \ac{soa} in each of the supported tasks (Fig.~\ref{fig:tasks:overview}). In parallel to this effort is the focus on improving and extending the \ac{fiw} dataset~\cite{robinson2016families, robinson2018visual, wang2017kinship}-- a large-scale, multi-task image set for kinship recognition.\footnote{\ac{fiw} project page, \href{https://web.northeastern.edu/smilelab/fiw/}{https://web.northeastern.edu/smilelab/fiw/}.} The size and scope of \ac{fiw} have proven to match the demand of modern-day, data-hungry deep networks~\cite{AdvNet, ertugrul2017will, gao2019will, li2017kinnet, wu2018kinship}. Nonetheless, there is room to grow in the quality, quantity, and even organization of the existing labels and evaluation protocols. We propose modern technologies to work with the data, along with plans to improve the scopes of the experiments (\ie more realistic settings). 
 


\begin{table*}[!t]
    \centering
    \caption{Counts for T-1: the number of unique pairs (\textbf{P}), families (\textbf{F}), and faces (\textbf{S})~\cite{robinson2020recognizing}.}
     \resizebox{\textwidth}{!}{%
    \begin{tabular}{p{.1in}r c c c c c c c c c c c c}
        & &\multicolumn{3}{c}{\cellcolor{blue!15}\textbf{{1$^{st}$ Generation}}} & \multicolumn{4}{c}{\cellcolor{red!15}\textbf{{2$^{nd}$ Generation}}}  & \multicolumn{4}{c}{\cellcolor{green!15}\textbf{{3$^{rd}$ Generation}}}\\

        & & \rot{\emph{\acs{bb}}} & \rot{\emph{\acs{ss}}} & \rot{\emph{\acs{sibs}}} & \rot{\emph{\acs{fd}}} & \rot{\emph{\acs{fs}}} & \rot{\emph{\acs{md}}} & \rot{\emph{\acs{ms}}} & \rot{\emph{\acs{gfgd}}} & \rot{\emph{\acs{gfgs}}} & \rot{\emph{\acs{gmgd}}} & \rot{\emph{\acs{gmgs}}}  & {\emph{Total}}\\ 
     \parbox[t]{2mm}{\multirow{3}{*}{\rotatebox[origin=c]{90}{train}}}&\textbf{P} & 991  & 1,029 &1,588 & 712 & 721& 736& 716 & 136 & 124 & 116 & 114 &6,983\\
    \multirow{3}{*}{} &\textbf{F}  &303 & 304 & 286 & 401 & 404 & 399 & 402 & 81 & 73&71 & 66 &2790\\
    \multirow{3}{*}{} &\textbf{S} &39,608& 27,844 & 35,337& 30,746  &46,583 & 29,778&  46,969& 2,003 &  2,097  &1,741 & 1,834  &264,540\\
 
     \rowcolor[HTML]{EFEFEF}
   &\textbf{P}  & 433 & 433 & 206& 220 & 261 & 200 & 234 & 53 & 48 & 56 & 42 & 2,186 \\
    
    \rowcolor[HTML]{EFEFEF}
   \parbox[t]{2mm}{\rotatebox[origin=c]{90}{val}}  &\textbf{F}  &74  & 57& 90 & 134& 135& 124& 130& 32& 29& 36&27 &868\\
    
  \rowcolor[HTML]{EFEFEF}   
   \multirow{-3}{*}{} &\textbf{S}  & 8,340 & 5,982 & 21,204& 7,575 &9,399&8,441 &7,587 & 762 &879 & 714 & 701 & 71,584\\
    
    \parbox[t]{2mm}{\multirow{3}{*}{\rotatebox[origin=c]{90}{test}}} &\textbf{P}  &  469& 469 & 217 & 202& 257 & 230 & 237 & 40 & 31 & 36 & 33&2,221 \\
    \multirow{3}{*}{} &\textbf{F}  & 149  & 150  & 89 & 126 & 133 & 136 & 132 & 22 & 21 & 20 & 22 & 1,190\\
    \multirow{3}{*}{} &\textbf{S}  & 3,459 &2,956 &967 &3,019&3,273&3,184& 2,660 &121&96&71&84&39,743
    \end{tabular}}\label{tbl:track1:counts} 
        \vspace{.2in}
\end{table*}

 Specifically, the \ac{fiw} dataset introduced imagery scraped \emph{in the wild}, with unconstrained family photos from the web. Although it is the \textbf{largest} and \textbf{most comprehensive} dataset of its kind, there still exist several concerns. For starters, the diversity of the data: the near 1,000 families are vast compared to other datasets, but not compared to the real world. Furthermore, still-faces in images are not the only target in multimedia that can infer kinship, the facial dynamics in videos and speech signals were shown to complement still-faces (\ie \ac{fiwmm}). Hence, future \ac{rfiw} will support the multimedia content. Lastly, the experimental settings should be tweaked and further bridge the gap between research and reality. 

Like in~\cite{robinson2020recognizing}, the 2021 \ac{rfiw} comprised three tasks depicted in Fig.~\ref{fig:tasks:overview}: (\textbf{T-1}) Kinship verification, (\textbf{T-2}) Tri-subject verification, and (\textbf{T-3}) Family member search \& retrieval.

The rest of the paper is organized as follows. First, we review the related (Section~\ref{sec:relatedworks}). A brief review of the data and task protocols follows (Section~\ref{sec:task:eval}). We then introduce the top methods of the challenge (Section~\ref{sec:submissions}). Lastly, we end with a discussion (Section~\ref{sec:discussion}) and a conclusion (Section~\ref{sec:conclusion}).

\section{Related Works}\label{sec:relatedworks}

Kinship understanding is a critical vision problem first published in the 2010 ICIP~\cite{fang2010towards}. As a human-centered visual learning problem, earlier works focused on low-level features of facial imagery. To combat challenges inherent in visual kinship verification (\eg variations from age), researchers incorporated mainstream learning approaches like transfer learning \cite{Xia201144,xia2012understanding}, and metric learning \cite{lu2014neighborhood,wang2017kinship}. More recently, advances in the evaluation protocols paved the way to pragmatic problem formulations (\eg tri-subject kinship verification \cite{qin2015tri} and family recognition~\cite{robinson2016families, robinson2018visual}). Furthermore, as a part of this \ac{rfiw} data challenge series, with its debut in last year's edition~\cite{robinson2020recognizing}, search and retrieval of missing family problems mimic a variety of practical use-cases.\footnote{\href{https://towardsdatascience.com/families-in-the-wild-track-iii-b5651999385e}{https://towardsdatascience.com/b5651999385e}} Thus, they further closed the gap from research to reality. Robinson~\etal analyze the evolution of the visual kinship problem domain over time, along with the various paradigms, \ac{soa}, and promising future directions in a recent survey~\cite{robinson2021survey} and dissertation~\cite{robinson2020automatic}. 

During the past decade, deep learning has been prominent across face-based vision systems~\cite{wang2020deep}. Initial visual kinship benchmarks (\eg KinWild \cite{lu2014neighborhood} and Family101 \cite{fang2013kinship}) had a profound impact in organizing and promoting the problem. However, such minimal data were insufficient to match the capacity needed to train deep models, with few exceptions (\eg in ~\cite{zhang12kinship}, smaller, parts-based models gained an incremental boost in performance compared with the pre-existing low-level methods). Furthermore, these earlier datasets mostly used faces from the same photos, such as color features~\cite{lopez2016comments} and then with the \emph{same photo} detectors~\cite{dawson2018same} claimed \ac{soa}, highlighting problems with constructing a verification set using face samples from the same photo.

The shortage of data quality and labels, and, hence, the absence of proper data distribution of the faces of families, motivated the release of the large-scale dataset \ac{fiw}~\cite{FIW}. Ever since, \ac{fiw} has supported various deep learning approaches~\cite{wu2018kinship, wei2019adversarial}, with some generative models to predict the appearances of family members~\cite{gao2019will, ozkan2018kinshipgan}. Even other multimodal views (\ie \ac{fiwmm}~\cite{robinson2021families}), and in tutorials at top-tier conferences (\ie ACM MM~\cite{robinson2018recognize} and CVPR\footnote{\href{https://web.northeastern.edu/smilelab/cvpr19_tutorial/}{https://web.northeastern.edu/smilelab/cvpr19\_tutorial/}}).

As a series of workshops and data challenges based on \ac{fiw}, \ac{rfiw} has been held in different venues over the past four years~\cite{robinson2017recognizing,robinson2020recognizing}. Plus, \ac{fiw} premiered in a Kaggle competition that attracted over 500 teams to make submissions.\footnote{\href{https://www.kaggle.com/c/recognizing-faces-in-the-wild}{https://www.kaggle.com/c/recognizing-faces-in-the-wild}} As part of the \ac{fg}, we hosted the fifth edition of \ac{rfiw} (\ie 2021 \ac{rfiw}). Hence, there is a continued effort to keep updating and promoting the \ac{fiw} dataset to push \ac{soa} and inspire researchers in the years to come.

\begin{table*}[!t]
    \centering
\caption {Averaged verification accuracy scores for \textbf{T-1}.}
     \resizebox{\textwidth}{!}{%
    \begin{tabular}{r c c c c c c c c c c c c}
        &\multicolumn{3}{c}{\cellcolor{blue!15}\textbf{{1$^{st}$ Generation}}} & \multicolumn{4}{c}{\cellcolor{red!15}\textbf{{2$^{nd}$ Generation}}}  & \multicolumn{4}{c}{\cellcolor{green!15}\textbf{{3$^{rd}$ Generation}}}\\
        
         \emph{Team} & \rot{\emph{\acs{bb}}} & \rot{\emph{\acs{ss}}} & \rot{\emph{\acs{sibs}}} & \rot{\emph{\acs{fd}}} & \rot{\emph{\acs{fs}}} & \rot{\emph{\acs{md}}} & \rot{\emph{\acs{ms}}} & \rot{\emph{\acs{gfgd}}} & \rot{\emph{\acs{gfgs}}} & \rot{\emph{\acs{gmgd}}} & \rot{\emph{\acs{gmgs}}}  & {\emph{Average}}\\ 
  Baseline~\cite{robinson2020recognizing} & 0.61 & 0.66 & 0.69 & 0.62 & 0.66 &0.71& 0.73 & {0.68} & 0.57 & 0.64 & 0.50 & 0.64\\
TeamCNU~\cite{id2new}&	\textbf{0.82}  &	\textbf{0.84}  &	\textbf{0.80}  &	0.76  &	\textbf{0.82}  &	\textbf{0.75}  &	\textbf{0.77}  &	\textbf{0.76}  &	0.71  &	0.75  &	0.59  &	\textbf{0.80}  \\
vuvko~\cite{id4}&	0.75  &	0.81  &	0.78  &	0.74  &	0.78  &	0.69  &	0.76  &	0.60  &	\textbf{0.80}  &	\textbf{0.80}  &	\textbf{0.77}  &	0.78  \\
nc2893~\cite{id1}&	0.76  &	0.78  &	0.75  &	0.74  &	0.70  &	0.67  &	0.70  &	0.59  &	0.79  &	0.79  &	0.75  &	0.77  \\
jh3450~\cite{id1}&	0.76  &	0.78  &	0.75  &	0.74  &	0.70  &	0.67  &	0.70  &	0.59  &	0.79  &	0.79  &	0.75  &	0.77  \\
paw2140~\cite{id1}&	0.75  &	0.78  &	0.76  &	0.74  &	0.68  &	0.69  &	0.72  &	0.59  &	0.78  &	0.79  &	0.75  &	0.77  \\
DeepBlueAI~\cite{id3}&	0.74  &	0.81  &	0.75  &	0.74  &	0.72  &	0.73  &	0.67  &	0.68  &	0.77  &	0.77  &	0.75  &	0.76  \\
ustc-nelslip~\cite{id6, id3new}&	0.76  &	0.82  &	0.75  &	0.75  &	0.79  &	0.69  &	0.76  &	0.67  &	0.75  &	0.74  &	0.72  &	0.76  \\
stefhoer~\cite{id2}&	0.77  &	0.80  &	0.77  &	\textbf{0.78}  &	0.70  &	0.73  &	0.64  &	0.60  &	0.66  &	0.65  &	0.76  &	0.74  \\
    \end{tabular}}\label{tab:benchmark:track1}
       \vspace{.2in}
\end{table*}

\section{Task Evaluations, Protocols, Benchmarks}\label{sec:task:eval}
We briefly introduce each task and refer readers to our earlier white paper for more details~\cite{robinson2020recognizing}. Also, see Fig.~\ref{fig:tasks:overview} for a visual depiction of each.

Historically, most focus has been on \textbf{T-1}~\cite{duan2017advnet, li2017kinnet, wang2017kinship, wu2018kinship}. Then, we introduced \textbf{T-2} and \textbf{T-3} in 2020~\cite{robinson2020recognizing}, for which several participants engaged~\cite{id2, id3, id4, id5, id6, id8, id9}. The three data splits are formed at the family level to enable multi-task solutions, along with the earlier need to keep the ground truth for a subset of the families from the public (\ie for blind testing). In other words, the three sets (\ie \emph{train}, \emph{val}, and \emph{test}) have the same families across all tasks. Specifically, 60\% of the families make up the \emph{train} set, while the remaining 40\% was split between \emph{val} and \emph{test}. Hence, the three sets are disjoint in family and identity, which remain consistent across the different tasks. 

Since the ground truth was released after the last \ac{rfiw}, it was fair to supply all data and labels. Teams were asked to only process the \emph{test} set when generating submissions, and any attempt to analyze or understand the \emph{test} pairs were prohibited. Also, outputs were scored on the server, for which we received and scored all submissions via Codalab (\ie \textbf{T-1}\footnote{\href{https://competitions.codalab.org/competitions/21843}{https://competitions.codalab.org/competitions/21843}}, \textbf{T-2}\footnote{\href{https://competitions.codalab.org/competitions/22117}{https://competitions.codalab.org/competitions/22117}}, and \textbf{T-3}\footnote{\href{https://competitions.codalab.org/competitions/22152}{https://competitions.codalab.org/competitions/22152}}).

All faces were encoded via Sphereface \ac{cnn}~\cite{Liu_2017_CVPR}  (\ie 512 D), with the pre-processing and training from the original work.\footnote{\href{https://github.com/wy1iu/sphereface}{https://github.com/wy1iu/sphereface}} Cosine similarity determined the closeness of pairing faces by comparing features $p_1$ and $p_2$, which is defined as
$\text{CS}(\pmb p_1, \pmb p_2) = \frac {\pmb p_1 \cdot \pmb p_2}{||\pmb p_1|| \cdot ||\pmb p_2||}$~\cite{nguyen2010cosine}.

\begin{table}[t]
\centering
\caption {Verification accuracy scores for \textbf{T-2}.}
\label{tab:benchmark:track2}
 \resizebox{.85\linewidth}{!}{%
\begin{tabular}{rccc}
  \textbf{Team}&\cellcolor{blue!15}\textbf{FMS} & \cellcolor{red!15}\textbf{FMD} & \textbf{Avg.} \\
  Baseline~\cite{robinson2020recognizing}& 0.68 & 0.68 & 0.68 \\ 
  TeamCNU~\cite{id2new}&	\textbf{0.86}  &	\textbf{0.82}  &	\textbf{0.84}  \\
    stefhoer~\cite{id2} & 0.74 & 0.72 & 0.73 \\
  DeepBlueAI~\cite{id3}  & 0.77 & 0.76 & 0.77 \\
 ustc-nelslip~\cite{id6, id3new}  & {0.80} & {0.78} & {0.79} \\
\end{tabular}
}
   \vspace{.1in}
\end{table}

\subsection{Kinship Verification (\textbf{T-1})}\label{sec:kinver}

To verify kinship is predicting whether a pair of individuals are blood relatives. In computer vision, we compare faces to classify the pairs as KIN or NON-KIN as true or false, respectively. This one-to-one view of kinship recognition typically assumes prior knowledge in the relationship type~\cite{robinson2018recognize}. Hence, relationship types are evaluated independently. With the introduction of \ac{fiw}, the number of face pairs and relationship types for kinship verification (\ie \textbf{T-1}) has significantly increased. Three sets of the data (\ie \emph{train}, \emph{val}, and \emph{test}) are partitioned for \ac{rfiw} (Table~\ref{tbl:track1:counts}). The {\emph test} set had an equal number of positive and negative pairs, and no family (and, hence, subject identity) overlaps between sets. As of 2020, the challenge began to support grandparent-grandchild types, \ie \ac{gfgd}, \ac{gfgs}, \ac{gmgd}, \ac{gmgs}. Due to insufficient counts across folds, along with an incredible bias skewed away from the few families that make up pairs across three generations, the great grandparent-great grandchild pairs of \ac{fiw} are omitted from \textbf{T-1} of \ac{rfiw}.

 \begin{figure}[b!]
    \centering
    \includegraphics[width =\linewidth]{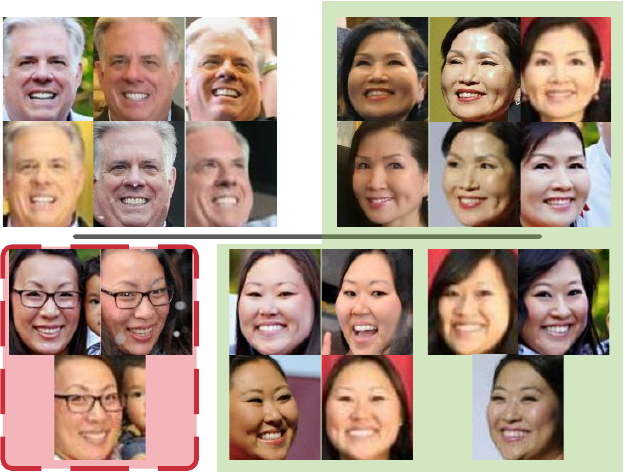}
    \caption{\diff{Sample family with mixed-race parents from \textbf{T-2} test set. \emph{Top row}: parents of the three daughters (\emph{bottom row}). Our baseline incorrectly labeled all triplets (\ie \emph{parent}-\emph{child} pairs) with the daughter shown left-most, while correctly classifying all pairs with the other two daughters (shown in the middle and right columns): the former scored less when compared to the mother, while the latter two daughters scored higher (\ie more similar) to their mother such that it outweighed the low similarity compared with the father (\ie all three daughters score low with their father).
    }}
    \label{fig:t2:qresults}
\end{figure}

Verification accuracy is used to evaluate the performance. Specifically, $\text{Accuracy}_j=\frac{\text{TP}_j + \text{TN}_j}{\text{N}_j}$,
where $j^{th}\in\{\text{all 11 relationship types}\}$. Then, the overall accuracy is the weighted sum. The threshold for positive and negative pairs was found by the value that maximizes the accuracy on the  \emph{val} set. Results are listed in Table~\ref{tab:benchmark:track1}.

\subsection{Tri-Subject Verification (\textbf{T-2})}\label{sec:trisubject}

Tri-Subject Verification focuses on a different view of kinship verification-- the goal is to decide if a child is related to a pair of parents. First introduced in~\cite{qin2015tri},  it makes a more realistic assumption, as knowing one parent often means the other potential parent(s) can be easily inferred.

Triplet pairs consist of Father ({F}) / Mother ({M}) - Child ({C}) ({FMC}) pairs, where the child {C} could be either a Son ({S}) or a Daughter ({D}) (\ie triplet pairs are {FMS} and {FMD}).

\begin{figure}[h!t]
\begin{center}
    
\includegraphics[width=\linewidth]{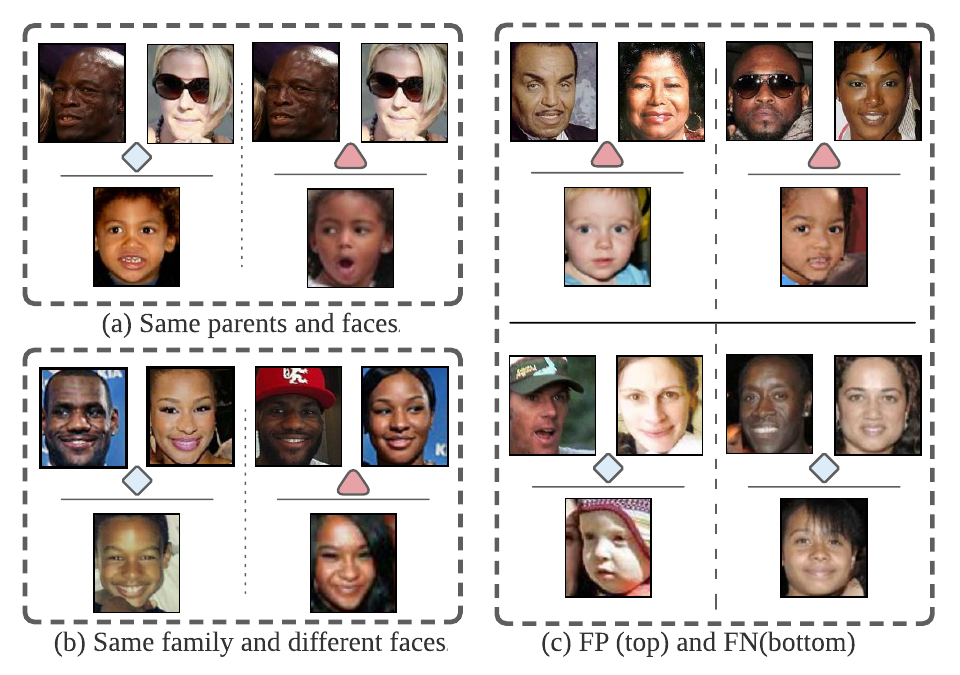} 
\end{center}
\vspace{3mm}
\caption{\diff{Triplets of \textbf{T-2} incorrectly predicted, with the parents above the child separated by a shape standing for the type of error, where false negatives (FN) and false positives (FP) are marked by \textcolor{cyan!100}{$\mathbf{\Diamond}$} and \textcolor{red!100}{$\mathbf{\bigtriangleup}$}, respectively. Exemplified are both error types for the same parent face pair, with wrong classifications marking the true son as NON-KIN and a random daughter as KIN (a). Besides the mother's images with sunglasses occlude the periocular region, other challenges seem to be in the mixed race, for the FP and FN samples shown, although on the wrong side of the decision boundary, are not far from the other (\ie similar scores). The next triplet pairs incorrectly classified are of the same parents' pairs (\ie different faces). Again, the son is falsely classified as NON-KIN, while a random daughter is an FN (b). The last set of sample triplets incorrectly classified show challenges posed by challenges in age (\ie FN triplet with a baby face), while the others are of mixed-race parents (c).}}
\label{fig:t2:samples}
\end{figure}

\begin{figure*}[h!t]
\begin{center}

\begin{subfigure}{.23\textwidth}
  \includegraphics[width=\linewidth]{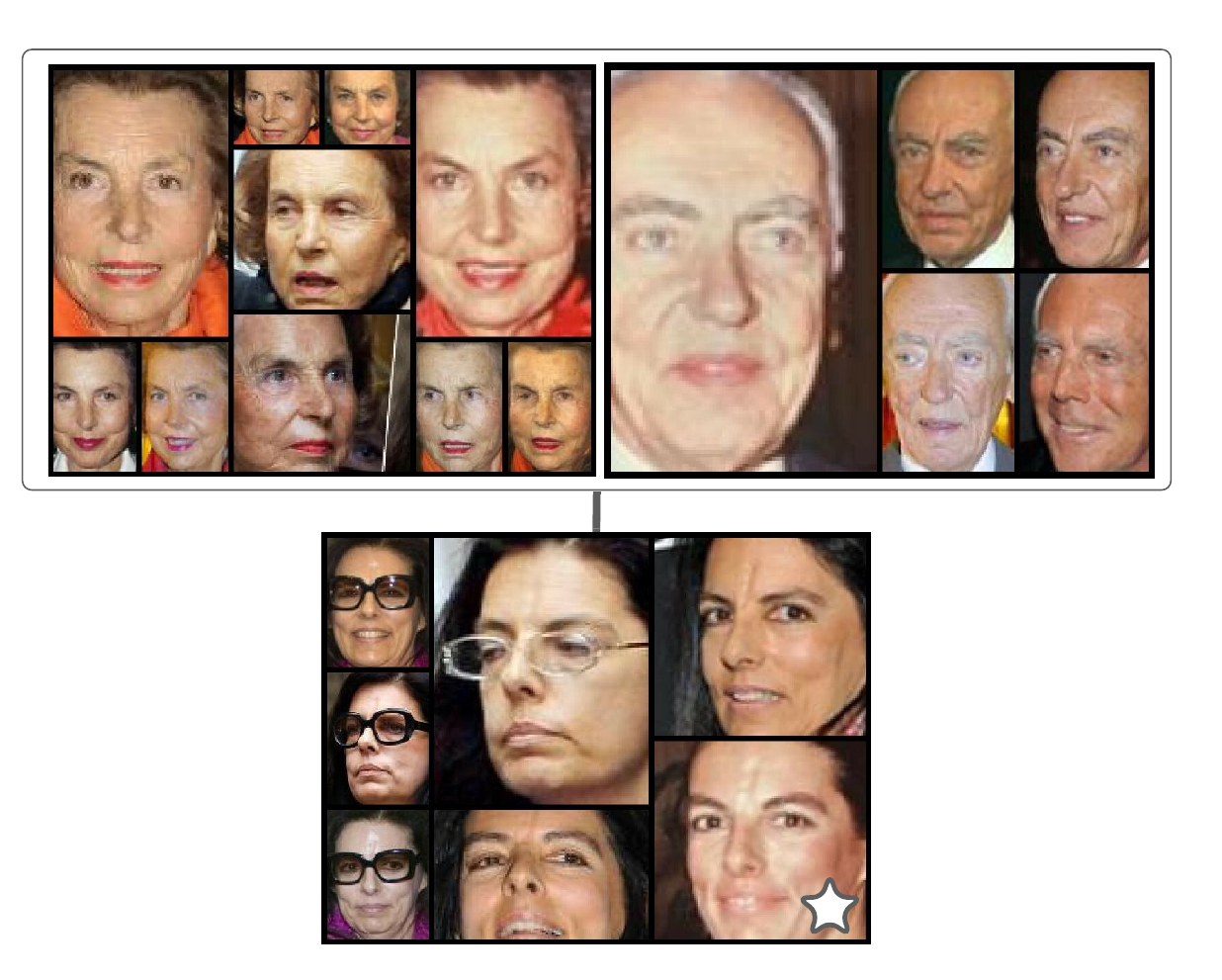}  
  \caption{Top scoring.}
  \label{fig:sub-first}
\end{subfigure}
\hspace{6mm}
\begin{subfigure}{.3\textwidth}
  \includegraphics[width=\linewidth]{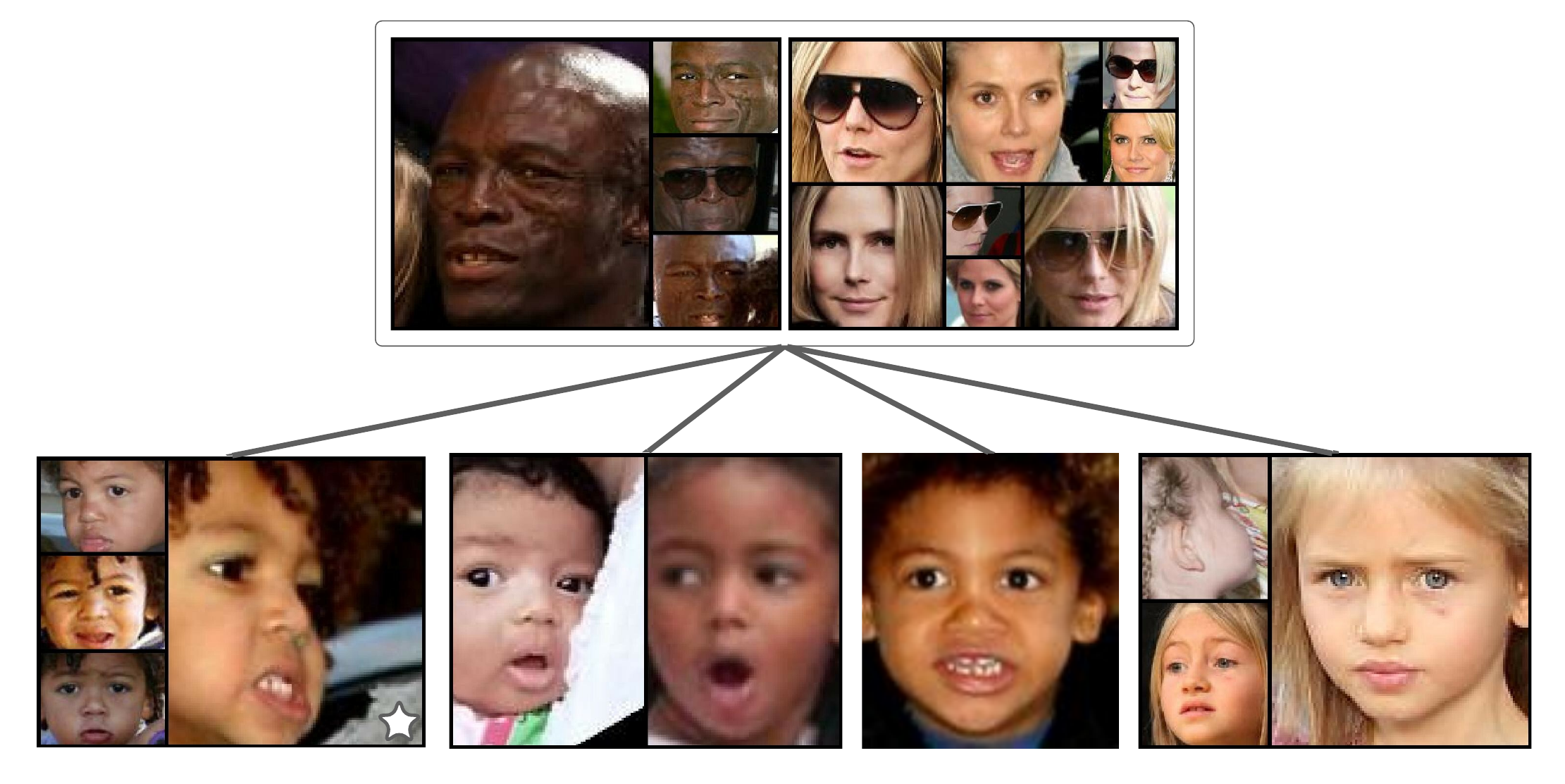}  
  \vspace{2mm}
  \caption{Runner-up (\ie $2^{nd}$ scoring).}
  \label{fig:sub-second}
\end{subfigure}
\hspace{3mm}
\begin{subfigure}{.3\textwidth}
  \includegraphics[width=\linewidth]{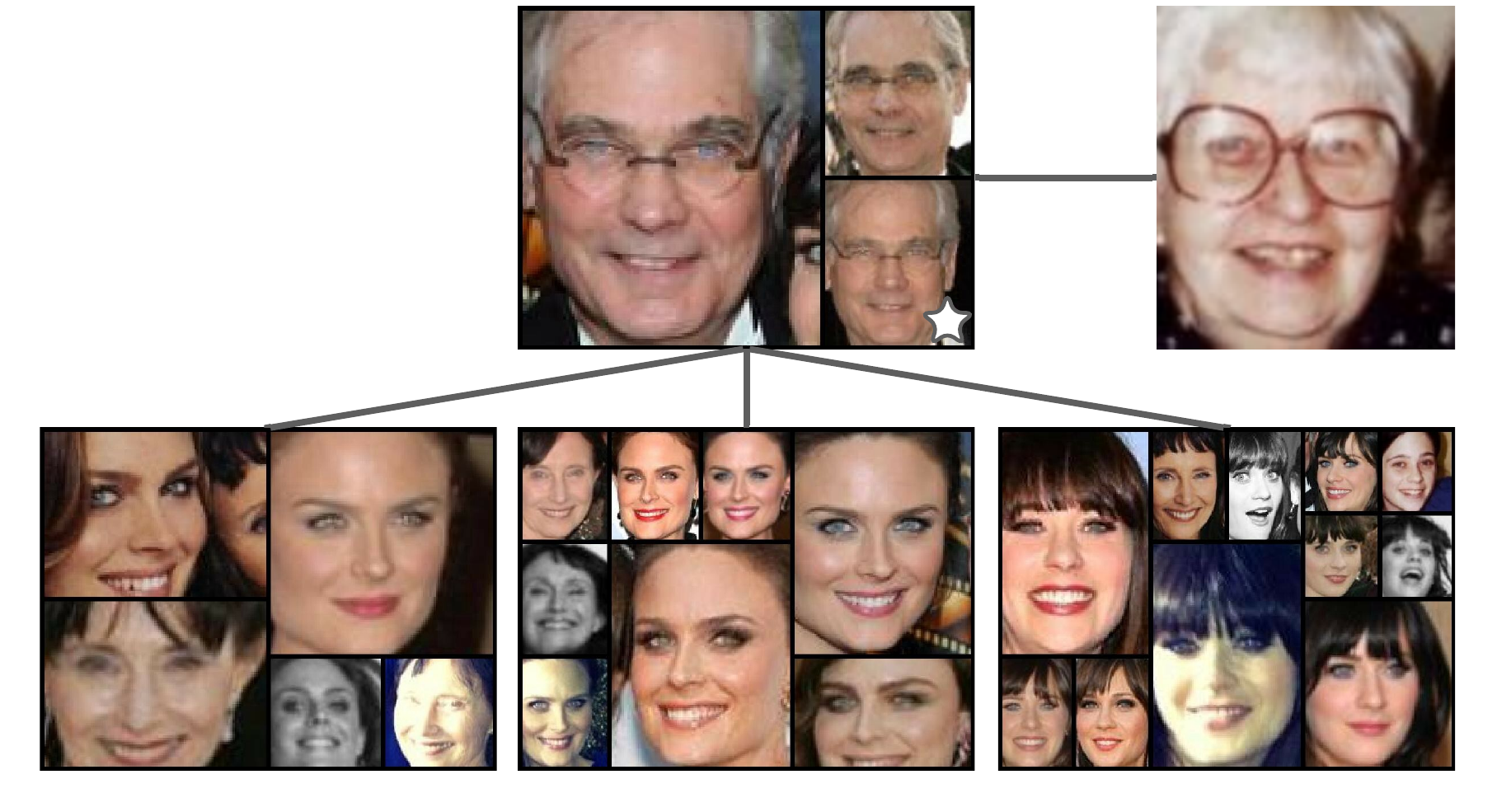}  
  \vspace{1.225mm}
  \caption{Lowest scoring.}
  \label{fig:sub-third}
\end{subfigure}
\end{center}
\caption{The top-ranking (a), runner-up (b), and last-place (c) families, on average, in \textbf{T-3}. The probe (\ie search query) displays a white star in the lower-left corners of the respective montage. The top two families (\ie (a) and (b)) are queries of children with both parents in the gallery, while the family with the lowest accuracy was a subject with children and a sibling (\ie a sister) present. Note that connections indicate blood relatives, where (a) and (b) the parents that share children (\ie the only blood connection shared is in the children), while the topmost of (c) does not include the mother of the children, for she is not a blood relative of the query (\ie who is a father and brother in the types of relationships in focus here).}
\label{fig:t3:familysamps}
\end{figure*}
Triplets were formed by first matching each mother-father pair with their biological children to form the list of positives. Then,  negative triplets were generated by shuffling the children in the positive list such that pairs of parents remained constant to yield the same number of negatives. Note that the number of possible negatives is far more than positives, so a pair of faces of a parent pair was used once and only once to produce a balanced list. Again, no family or subject identity overlaps between sets: the same families make up the \emph{train}, \emph{val}, and \emph{test} sets across the tasks. 

Verification accuracy WAS first calculated per triplet-pair type (\ie FMD and FMS), and then averaged via the weighted sum. A score was assigned to each triplet $(F_i, M_i, C_i)$ in the \emph{val} and \emph{test} sets using the formula $ \text{score}_{i} =  \text{mean}(\cos{(F_i, C_i)}, \cos{(M_i, C_i)})$,
where $F_i$, $M_i$, and $C_i$ are the feature vectors of the $i$-th triplet. 
Scores were compared to a threshold $\gamma$ to infer a label (\ie KIN if the score surpasses the threshold; else, NON-KIN). 
The threshold was found experimentally on the \emph{val} set. The threshold was applied to the \emph{test} (Table~\ref{tab:benchmark:track2}).

\subsection{Search and Retrieval (\textbf{T-3})}\label{sec:search}
As a search cue, kinship information can improve conventional FR search systems as prior knowledge for mining social or ancestral relationships in industries like \textit{Ancestry.com}. However, the task is most related to missing persons. Thus, we pose \textbf{T-3} as a set-based paradigm. For this, we imitate template-based evaluations on the \textit{probe} side but with a \textit{gallery} of faces~\cite{whitelam2017iarpa}. Furthermore, the goal is to find relatives of search subjects (\ie subject-level \textit{probes}) in a search pool (\ie face-level \textit{gallery}).

The protocol of \textbf{T-3} could be used to find parents and other relatives of unknown, missing children. The \textit{gallery} has 31,787 facial images from 190 families in the \textit{test} set. The inputs are media for a subject (\ie \textit{probes}), and outputs are ranked lists of all faces in the \textit{gallery}. The number of relatives varies for each subject, ranging anywhere from 1 to 20+. Furthermore, probes have one-to-many samples-- the means of fusing feature sets (\ie a \textit{probe's} media) is an open research question~\cite{zhaomulti}. This \textit{many-to-many} task is currently set up in closed form (\ie every \textit{probe} has relative(s) in the \textit{gallery}).

For each of the $N$ test \emph{probes} (\ie family $f$), the average precision (AP) is calculated: 
$$\text{AP}(f)=\frac{1}{P_F}\sum^{P_F}_{tp=1}\text{Prec}(tp)=\frac{1}{P_F}\sum^{P_F}_{tp=1}\frac{tp}{rank(tp)},$$
where ${P_F}$ is the number of \acp{tp} for the $f$-th family. The average AP (\ie \ac{map}) is then reported:
$\ac{map} = \frac{1}{N}\sum^{N}_{f=1}AP(f)$. Finally, Rank@5 is reported: the average number of \emph{probes} returned at least one \ac{tp} in the top five gallery faces. The baseline, along with others on the scoreboard, is shown in Table \ref{tbl:t3:benchmarks}.

\begin{table}[t!]
	\centering
	\caption{Performance ratings for \textbf{T-3}.}
	 \resizebox{\linewidth}{!}{%
\begin{tabular}{rccc}
	      \textbf{Team} &\cellcolor{blue!7}\textbf{Average} &\cellcolor{blue!10}\textbf{mAP} & \cellcolor{blue!15}\textbf{Rank@5} \\
	      ustc-nelslip~\cite{id6, id3new} & 0.35 & 0.15 & 0.54	\\
		  Baseline~\cite{robinson2020recognizing} & 0.28 & 0.11 & 0.45	\\
		  TeamCNU~\cite{id2new} &	\textbf{0.40}  &	\textbf{0.21}  &	\textbf{0.60}  \\
		  HCMUS notweeb~\cite{id9} & 0.17& 0.07 & 0.28	\\
		  DeepBlueAI~\cite{id3} & 0.19 & 0.06 & 0.32	\\
		  vuvko~\cite{id4} & 0.39 & {0.18} & \textbf{0.60}	\\
	\end{tabular}}
	\label{tbl:t3:benchmarks}
\end{table}

\section{Summary of submissions}\label{sec:submissions}
New solutions published as part of the 2021 \ac{rfiw} challenge in the \ac{fg} proceedings are introduced. Readers are referred to the paper references for additional details.

\paragraph{TeamCNU} proposed a contrastive learning framework to tackle and lead all three tasks~\cite{id2new}. The core idea of their solution is that self-supervised contrastive learning can help to learn powerful representations for different downstream tasks such as three tracks in RFIW. Following the 2020-RFIW's winning team Vuvko~\cite{id4}, ArcFace (\ie a pre-trained ResNet-101) was used to encode raw faces, and then a \ac{mlp} was attached to obtain the low-dimensional feature pair for computing the contrastive loss~\cite{chen2020simple}. To output the kinship verification results for all three tasks, it only needs to remove the MLP layers and take the mid-level features extracted by the Siamese backbone to compute the similarity score. Then, a predefined threshold is naturally applied to select the positive pairs. It surprisingly outperforms all the former SOTA methods on most of the tracks in the RFIW challenge.

\paragraph{paw2140} used ensemble learning on both data and networks levels~\cite{id1}. Specifically, this team applied data augmentation to obtain more representative input data, like random rotation in angle, minor crops, horizontal flips, and channel-wise transformation. Moreover, the authors used multiple duplicate data samples with different augmentations as the ensemble inputs for testing. The network ensemble has employed multiple structures such as ResNet50~\cite{he2016deep}, FaceNet~\cite{schroff2015facenet}, VGGFace~\cite{parkhi2015deep}, and SENet50~\cite{hu2018squeeze} as the backbones where the features from different backbones fused in multiple ways. The Hadamard product, squared, and absolute value difference of the pairs of features are concatenated for the high-level similarity quantization. Moreover, there is the ensemble across different splits of the training data with \emph{k} folds. For each instance, the final \emph{4*k} ensemble member networks, trained with k different splits and four Siamese networks for each split, are applied for the prediction. To further boost the performance, they take the program synthesis based on the OpenAI's Codex~\cite{chen2021evaluating} to automatically generate variants of networks for the ensemble.

\paragraph{ustc-nelslip} used a Siamese neural network designed for all three tasks~\cite{id3new}. For the \emph{one-vs-one} kinship verification (\textbf{T-1}), the team takes a two-branch deep Siamese neural network with the enhanced feature fusion (\ie concatenation of squared difference, the difference of squared features, and dot product). To address the \emph{two-vs-one} verification (\textbf{T-2}), they proposed a pair of deep Siamese neural networks, each included four branches. There is only one branch network for the child data shared by the two Siamese networks. Moreover, the other two branches are for mother and father images, respectively. Team ustc-nelslip introduced the feature fusion similarity and cosine similarity for measurement to obtain the rank of similar images of a query (\textbf{T-3}). Finally, the obtained similarity scores of mother-children and father-children are weighted to generate the parent-child similarity score for kinship verification.

\begin{figure*}[t!]
    \begin{subfigure}{.3\linewidth}
    \scriptsize
\begin{python}
import fiftyone as fo
import fiftyone.zoo as foz

"""
download and prepare test data and labels
"""
dataset = \
  foz.load_zoo_dataset("fiw", split="test")

"""
launch desktop app to explore test faces
"""
session = fo.launch_app(dataset)
\end{python}

    \end{subfigure}
    \hspace{.4in}
\begin{subfigure}{.65\linewidth}
    \centering
    \includegraphics[width =\linewidth]{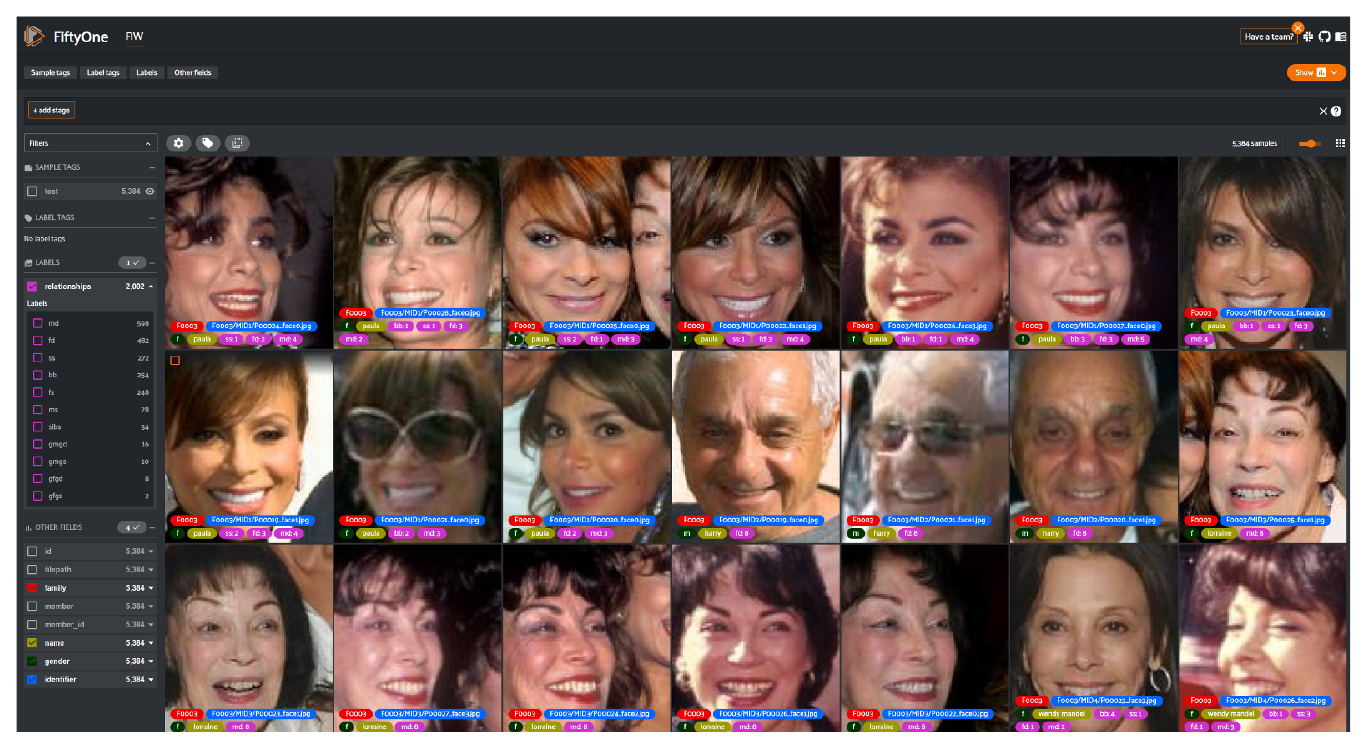}

        \end{subfigure}
            \caption{\acs{fiw} is now available as a part of fiftyone's datazoo. Fiftyone is an open-source tool for dataset curating and model analysis via an easy-to-use Python interface and application (\href{https://fiftyone.ai}{https://fiftyone.ai}). We aim for simpler access, built-in data exploration features, and to inspire community contributions through engaging. Hence, the small code snippet (left) launches the application, as shown (right). The left-side of the application window are filters, allowing the user to view different relationship types easily (\eg BB, SS, FD, MD) and filter families. Below the filters are check-boxes corresponding to label displays (\eg gender, image name, family label) - check to toggle metadata shown as colored labels at the bottom of each face. We chose fiftyone for various reasons, with it being extendable an essential spec: different labels and metadata are easy to add, manage, and extend. This incentive will support all tasks and, eventually, the multimedia variant \acs{fiwmm}~\cite{robinson2021families}. \ac{fiw} sets (\ie \emph{train}, \emph{val}, \emph{test}) are obtainable independently or all together. For instance, we expect analysis capabilities to be enhanced and the tracking of data bugs for an improved version rollout. Best viewed electronically.
            }
    \label{fig:fiftyone}
\end{figure*}

\section{Discussion}\label{sec:discussion}
\diff{Focusing on \textbf{T-2}, by not only depicting a shortcoming of the existing methods but also the potential for more advanced systems compared to models of \textbf{T-1}, let us use the sample family in Fig.~\ref{fig:t2:qresults} to examine the common challenge of mixed-race pairs. As described, the \emph{test} set is split by family, which stays constant for all tasks. Hence, the pair-list of \emph{T-2} is a subset of \emph{T-1} (\ie \emph{parent-child} pairs with children with both parents present). For instance, if son $S\in FS\And MS-> FMS$. Hence, the family shown is of type $FMD$, where all face pairs of $FD$ were falsely predicted as NON-KIN, while all $MD$ pairs were correctly inferred. As for the $FMD$ triplets, all pairs with one of their daughters were incorrect, while all others involving the other two daughters were correct. Hence, the information is there for the correct decision on all triplets, unlike with pairs of \textbf{T-1} (\ie all $FD$ were incorrect), but only two of three daughters look enough like the mother to offset the dissimilarity between the children and father.  A robust fusion scheme is a promising direction that could improve such cases (\ie mixed-race parents), especially considering how more common that is in modern-day families. 

Additionally, as depicted in Fig-\ref{fig:t2:samples}, we visually explore the common errors of \textbf{T-2}. For true and false pairs marked incorrectly classified, hard samples often include mixed-race parents. On average, the SOA \emph{parent-child} verification system is best with both parents present (\ie $\text{mAP}(FD,FS,MS,MD)<\text{mAP}(FMD,FMS)$. Hence, there is great promise in improving the fusion. We also can see, as usual, that a hard sample is of a baby's face (Fig.~\ref{fig:c}).
}

To explore easy and challenging cases of \textbf{T-3}, we took the average across the baseline and two submissions of this \ac{rfiw}. The top two queries in \ac{map} are of children, for which both parents are present in the gallery. The last-place query is a subject with no parents present in the query (Fig.~\ref{fig:t3:familysamps}). It is interesting to note that the parents of the second family are inter-racial: as expected, children tend to inherit features from both parents, on average.  

A challenge in facial recognition problems is age: it is typically more difficult when compared to a baby's face or an elder. Age variations are especially challenging in face-based visual kinship recognition and to the degree earlier works base the underlying motivation (\ie the story) about age. For instance, Wang~\etal viewed the problem as imbalanced data solved by augmenting data by transforming faces at age intervals~\cite{wang2018cross}. Note that the faces of the last place query are all at an older age (Fig.~\ref{fig:sub-third}). All blood relatives present in the gallery are the opposite sex (\ie one sister and three daughters), while both the top-scoring (Fig.~\ref{fig:sub-first}) and the runner-up (Fig.~\ref{fig:sub-second}) have at least half of their known (\ie present) relatives of the same sex with most of the faces.   

\ac{fiw} was added to fiftyone's datazoo~\cite{moore2020fiftyone} - accessing and exploring the data made efficient. Additionally, we plan to release evaluations per Voxel51's Python API (Fig.~\ref{fig:fiftyone}). \ac{fiw} is more accessible and extendable as we work to incorporate improved protocols, data quality, and MM in future \acp{rfiw} - whether raw faces, faces with predicted landmark (\ie generated via~\cite{robinson2019laplace}), aligned faces, or corresponding face encodings, a variety of versioned data will be easy to access). We also plan to incorporate our \ac{bfw}~\cite{robinson2020face,robinson2021balancing} similarly, which will aid much-needed studies of bias in the \ac{fiw} dataset.

\acresetall
\section{Conclusion}\label{sec:conclusion}
Another year of \ac{rfiw} in conjunction with the 2021 \ac{fg}, \ac{soa} in kin-based vision models continue to improve across all three tasks currently supported by the \ac{fiw} dataset. \emph{TeamCNU} topped the score charts via a contrastive learning framework geared to learn better representation for comparing faces in all three kinship recognition tasks. We are improving the existing data and settings by working to make \ac{fiw} more accessible while also bringing in the multimedia data in the benchmarks. Baseline code at \href{https://github.com/visionjo/pykinship}{github.com/visionjo/pykinship}.

{         
\bibliographystyle{ieee}

\scriptsize
\balance
\bibliography{rfiw2020} 

}
\end{document}